\pdfoutput=1

\documentclass[11pt]{article}

\usepackage{acl}

\usepackage{times}
\usepackage{ulem}
\usepackage{enumitem}
\usepackage{latexsym}
\usepackage{graphicx}
\usepackage{booktabs}
\usepackage{multirow}
\usepackage{amsmath}
\usepackage{amsfonts}
\usepackage[caption=false]{subfig}
\usepackage[T1]{fontenc}

\usepackage[utf8]{inputenc}
\usepackage{float}
\usepackage{microtype}

\usepackage{inconsolata}
\usepackage{dsfont}

\usepackage{CJKutf8} 

\title{Mix-Initiative Response Generation with Dynamic Prefix Tuning}

\author{Yuxiang Nie${}^{1}$\thanks{~~Work was done during an internship at SMU.},~~Heyan Huang${}^{1}$,~~Xian-Ling Mao${}^{1}$,~~Lizi Liao${}^{2}$ \\
        ${}^{1}$School of Computer Science and Technology, Beijing Institute of Technology\\
        ${}^{2}$ Singapore Management University \\
        \texttt{\{nieyx,hhy63,maoxl\}@bit.edu.cn,lzliao@smu.edu.sg}}

\begin{document}
\maketitle
\begin{abstract}
Mixed initiative serves as one of the key factors in controlling conversation directions. For a speaker, responding passively or leading proactively would result in rather different responses. However, most dialogue systems focus on training a holistic response generation model without any distinction among different initiatives. It leads to the cross-contamination problem, where the model confuses different initiatives and generates inappropriate responses. Moreover, obtaining plenty of human annotations for initiative labels can be expensive.
To address this issue, we propose a general mix-Initiative Dynamic Prefix Tuning framework (IDPT) to decouple different initiatives from the generation model, which learns initiative-aware prefixes in both \textit{supervised} and \textit{unsupervised} settings.
Specifically, IDPT decouples initiative factors into different prefix parameters and uses the attention mechanism to adjust the selection of initiatives in guiding generation dynamically. The prefix parameters can be tuned towards accurate initiative prediction as well as mix-initiative response generation.
Extensive experiments on two public dialogue datasets show that the proposed IDPT outperforms previous baselines on both automatic metrics and human evaluations. It also manages to generate appropriate responses with manipulated initiatives.\footnote{Our code and data are released at \url{https://github.com/JerrryNie/Mix-Initiative-Dialogue}.}
\end{abstract}

\section{Introduction}

Dialogue systems crept into our lives in various ways, such as social chatbots \citep{zheng2023building}, conversational recommenders \citep{dao2023reinforced,wu2023state}, task-specific assistants \citep{liao2018knowledge,qin2023end} and so on. Mixed initiative plays an important role in forming responses in these systems. For example, in a dialogue session, the user and the system can take turns to lead the conversation. The user could lead the conversation by introducing new topics, providing new requirements or negating the suggestions, while the system would also take initiative by actively asking questions or recommending items to the user.

\begin{figure}[t]
\centering
\vspace{-0.1cm}
\includegraphics[width=\linewidth]{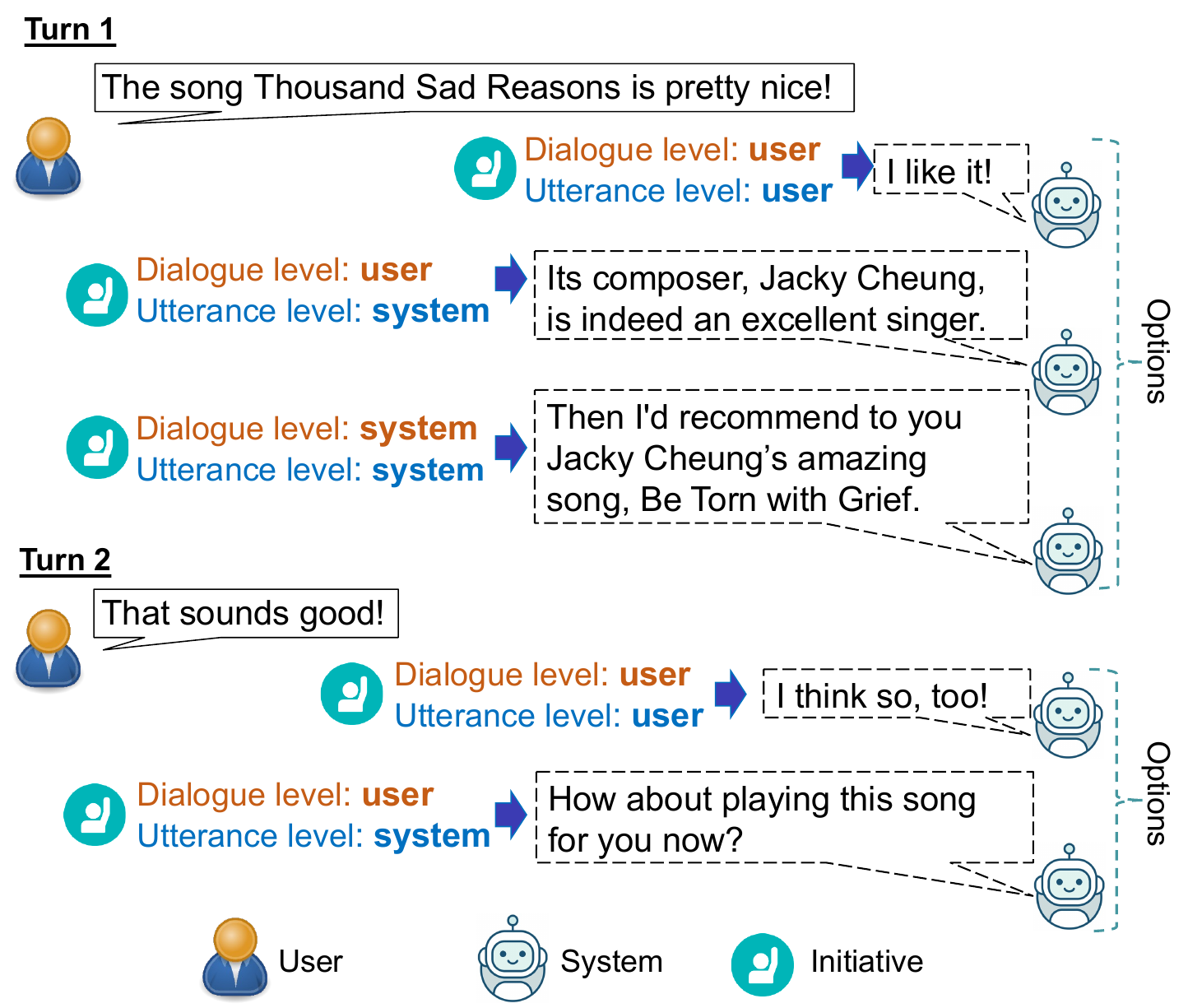}
\vspace{-0.5cm}
\caption{\label{fig:illustration} An illustration of initiative-aware response generation. Given a dialogue context, the possible responses vary with initiatives.}
\vspace{-0.3cm}
\end{figure}

However, current dialogue systems pay much attention to training a \textit{holistic} response generation model, neglecting to distinguish responses in terms of \textit{initiative}, which could lead to the cross-contamination problem \citep{liao2023proactive, liao2023p}. As shown in Figure \ref{fig:illustration}, when the user says, ``\textit{The song Thousand Sad Reasons is pretty nice!}'', the system could merely follow the topic, hence an appropriate response would simply be: ``\textit{I like it!}''. Yet, the system would also take control of the dialogue and recommend a new item. Then the response will change to something like: ``\textit{Then I'd recommend to you Jacky Cheung’s amazing song, Be Torn with Grief}''. Suppose we feed these to train a \textit{holistic} response generation model without distinction, what we get will mostly likely be an `averaged' response, which would largely affect the accuracy and flexibility of responses hence hurt human satisfaction.

There are indeed ways to introduce these affecting factors into the response generation model. For example, in empathetic dialogue systems, varied emotions also result in different responses. Hence, one possible solution is to directly adapt these empathetic response generation models such as \citep{majumder2020mime,chen2022emphi,qian-etal-2022-controllable}. However, there are differences in nature between empathy and initiative, which makes the adaption not appropriate. Firstly, emotional tone is relatively consistent between dialogue participants and the change of emotional tone is less frequent. Secondly, for responses with different emotions, there are only limited variations of patterns in responses \cite{chen2022emphi}. But for responses with different initiatives, the responses vary largely. Some researchers \citep{chen-etal-2023-controllable} design various strategies to prompt large language models (LLMs) for initiative-aware response generation. However, this kind of method requires deep understanding of each specific dataset to design effective prompts, which is of heavy workload. Another possible solution is to fine-tune separate pre-trained language models for each initiative. However, a well-performed response generation model requires a large number of responses with initiative labels, but obtaining these labels is costly and even impractical. Moreover,
It also takes up a large amount of disk space to save each fine-tuned model for each initiative. Therefore, motivated by \citet{li2021prefix}, we study a prefix-based method to inject these initiative factors into separate network parameters for better generality and disk efficiency.

To this end, we propose mix-Initiative Dynamic Prefix Tuning (IDPT), a general and disk-efficient model to decouple different initiatives from the response generation model. IDPT can work in both supervised and unsupervised settings, with limited or no initiative labels in the response training data. Specifically, 
IDPT tends to push initiative factors into different prefix parameters, while leaves the main generation task to the pre-trained language model. In \textit{supervised} setting (low-data) with initiative labels, it recognizes initiatives and applies hard attention over initiative prefixes for response generation. In \textit{unsupervised} setting without labels, it applies soft attention to consider various initiative prefixes for response generation.
Through training, IDPT learns to select the proper prefix parameters to combine with the pre-trained language model for mix-initiative response generation.

To sum up, our contributions are as follows:
\begin{itemize}[nosep]
    \item We propose to study mixed initiative in neural response generation models, which has the potential to improve the accuracy and flexibility of generation.
    
    \item We propose a general prefix-based neural response generation model which dynamically adjusts the selection and contribution of various initiatives during the generation process when there is limited or no initiative labels.

    \item Extensive experiments on two datasets show that the proposed IDPT outperforms baselines on both automatic metrics and human evaluations. IDPT also manages to generate proper responses with specified initiatives.
    
\end{itemize}

\section{Related Works}
\paragraph{Response Generation in Dialogue System}
Response generation serves as an important role in information exchange between humans and the dialogue system. Traditionally, recurrent neural network (RNN) \citep{rumelhart1985learning} based models are used in response generation, for Context-Sensitive Generation \citep{sordoni2015neural}, persona dialogue \citep{li2016persona}, empathetic dialogue \citep{chen2022emphi}, etc. Recently, with the help of the pre-trained language model like GPT-2 \citep{radfordlanguage}, the quality of the generated responses has improved much \cite{ye2022structured,ye2022reflecting}. \citet{hosseini2020simple} convert all sub-tasks in the task-oriented dialogue system into a single sequence prediction task so that response generation can benefit from task-related structural information. \citet{majumder2020mime} propose an empathetic response generation model via recognizing and mimicking the varied emotions from human utterances. \citet{chen2019towards} enhance the interaction between recommendation and response generation by using a relational graph convolutional network over a knowledge graph, while \citet{zhou2020improving} use knowledge-graph-based semantic fusion. 
However, these methods ignore the initiative in the generated response, which is essential in controlling the direction of the dialogues.

\paragraph{Initiative in Dialogue System}
The initiative \citep{walker1990mixed} is an important factor in driving the direction of a conversation.
\citet{DBLP:conf/sigir/VakulenkoKR20} analyze the relations among different conversation datasets in terms of the initiative. \citet{meng2021initiative} propose a self-supervised method to incorporate initiative in the knowledge selection phase of a dialogue system. \citet{chen-etal-2023-controllable}  prompt large language models (LLMs) for initiative-aware response generation. However, these methods either neglect the essential influence of the initiative on generating responses or use complex prompts for initiative-aware response generation.

\paragraph{Prompt-based Learning} Prompt-based learning \citep{liu2021pre} aims to achieve a text prediction task via directly modeling the properties of a pre-trained language model. \citet{yin2019benchmarking} reformulate the text to be classified as a slot-filling prompt and use a pre-trained Language model to fill this slot for classification. \citet{chen2021adaprompt} regard the relation extraction task as a fixed-prompt language model tuning task. \citet{li2021prefix} propose a tunable task-specific prefix to control a fixed language model and generate the text.

\begin{figure*}[t]
\centering
\includegraphics[width=0.9\linewidth]{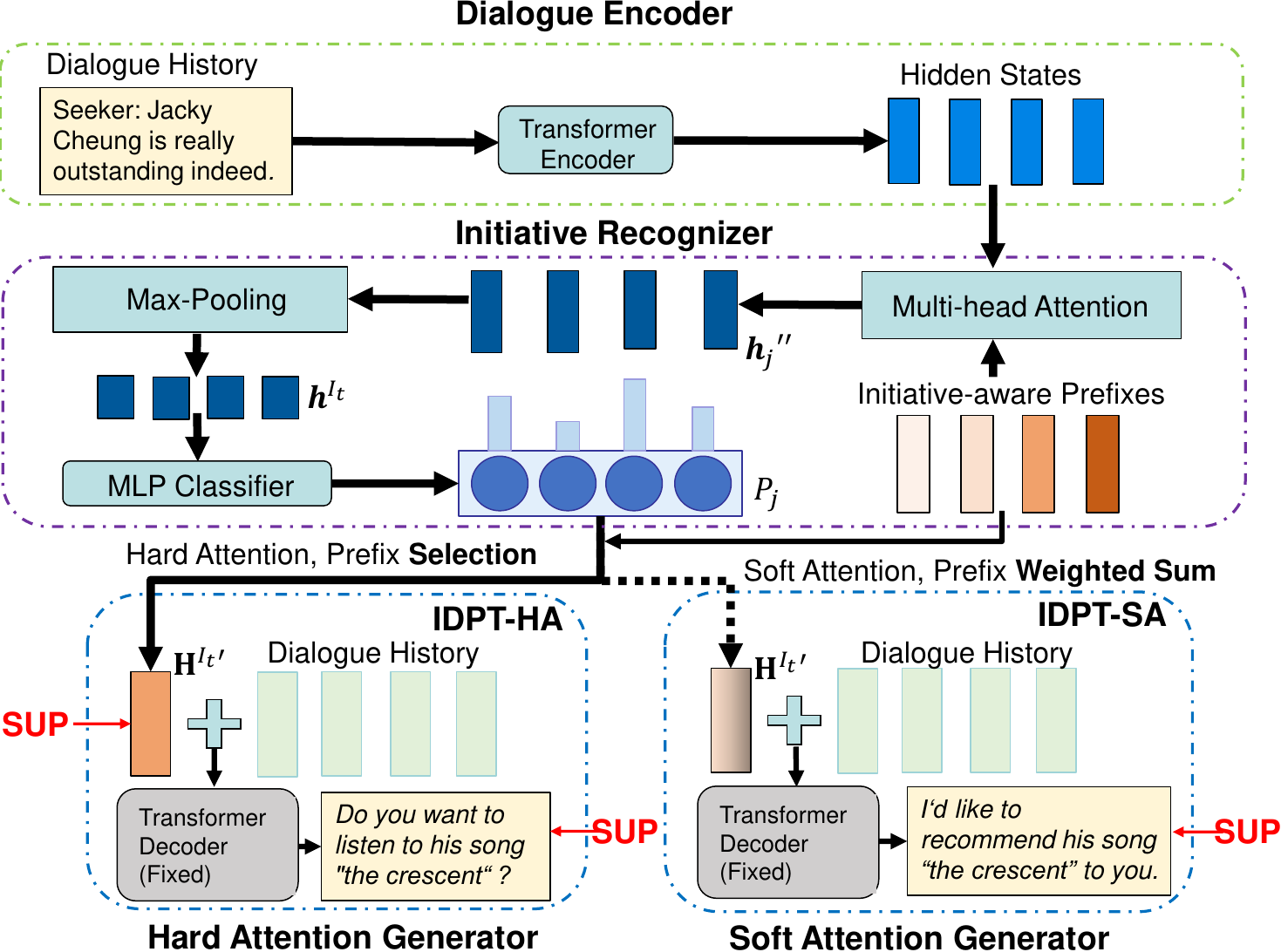}
\caption{\label{fig:model} Architecture of the proposed IDPT. It consists of a dialogue encoder, an initiative recognizer and a prefix-based response generator which has two realization versions (Hard for \textit{supervised} \textit{v.s.} Soft for \textit{unsupervised}). 
}
\end{figure*}

\section{Mix-Initiative Response Generation}

\subsection{Preliminary}

\paragraph{Initiative Classes.}
\label{sec:initiative_classes}
 Initiatives are a key concept in dialogue systems, but they are often conceptualized differently by different researchers \citep{chu1997tracking,hardy2021effective}. In this paper, we propose a hierarchical model of initiatives, which distinguishes between dialogue-level and utterance-level initiatives. Furthermore, we assume that each level of initiative can be either user-initiated or system-initiated, resulting in four possible categories of initiatives\footnote{Note that our model is applicable to both the user and the system roles, but for simplicity, we will focus on the system role in this paper.}.
 To make it clear, we organize these four initiatives in Table \ref{tab:init_class}. For example,  the initiative in the upper left of Table \ref{tab:init_class} corresponds to the initiative label of the response ``\textit{Then I'd recommend to you Jacky Cheung's amazing song, Be Torn with Grief}'' in Figure \ref{fig:illustration}. It means that for the current system response, the system takes control at the dialogue level as well as the utterance level.
 \begin{table}[!htp] 
    \centering
    \small
    \begin{tabular}{c|c|c}    
        \hline
         &UI: System& UI: User\\
         \hline
        DI: System& 3 & 2\\
        \hline
        DI: Utterance& 1 & 0\\
        \hline
    \end{tabular}
    \caption{\label{tab:init_class} Dialogue-level and utterance-level initiatives. ``DI'' means dialogue-level initiative and ``UI'' means utterance-level initiative. ``0''``1''``2''``3'' are label indices used for classification (Section \ref{sec:initiative_recognizer}). Thus, initiative label number is 4.} 
\end{table}

\paragraph{Task Formulation.}
Assume that we have a conversation $C=\{(X_t,Y_t)\}_{t=1}^{T}$ with $T$ turns, a knowledge text $K$ describing the background of the conversation as well as $T$ initiative labels $\{I_t\}_{t=1}^{T},I_t\in \{0,1,...,N\}$ corresponding to each $Y_t$, where we organize initiative states into $N$ classes as shown in Table \ref{tab:init_class}. In each turn $t$, given the knowledge $K$ and the dialogue history, which is $\{(X_t,Y_t)\}_{t=1}^{t - 1}$ and the user utterance $X_t$, the aim is to predict the proper initiative $I_t$ for response and produce the mix-initiative response $Y_t=(y_{t,1},y_{t,2},...,y_{t,|Y_{t}|})$ with $|Y_{t}|$ tokens.

\subsection{The IDPT Model}

The overview architecture of IDPT is shown in Figure \ref{fig:model}. It is composed of three modules: a dialogue encoder, an initiative recognizer and a prefix-based response generator, where the generator has two versions of realizations (Hard \textit{v.s.} Soft). The Hard version (IDPT-HA) is for the \textit{supervised} setting where a few initiative labels are available during training. It uses the initiative recognizer to predict a distribution $P_j$ over the possible initiative labels and selects one of the corresponding initiative-aware prefixes to generate a response. The Soft version (IDPT-SA) is designed for the \textit{zero-shot} setting where no initiative lable is available during training. It applies a soft attention mechanism to combine all the initiative-aware prefixes with weights derived from predicted $P_j$ and generates a response with the mixed prefix. The initiative recognizer and the response generator are trained with supervision signals (``\textit{SUP}'') only from the dialogue response.

\subsubsection{Dialogue Encoder}
Specifically, in the dialogue encoder, dialogue history\footnote{If there is no further explanation, ``dialogue history'' denotes the concatenated text of background knowledge and the dialogue history.} is encoded by a Transformer encoder (Encoder) model to obtain its contextualized representation.
To model the dialogue history text for initiative classification, IDPT first concatenates the knowledge text $K$ and the dialogue history into a sequence $\{S_{i}\}_{i=1}^{|S|}$, where $|S|$ is the length of dialogue history together with knowledge text. Then, $S$ is fed into an Encoder model:
\begin{equation}
\mathbf{H} = \text{Encoder}(S)~,
\end{equation}
where $\mathbf{H}\in\mathbb{R}^{|S|\times h}$ is the contextualized representation of the input $S$, $h$ is the hidden size of $\mathbf{H}$.

\subsubsection{Initiative Recognizer}
\label{sec:initiative_recognizer}
In the initiative recognizer as shown in the middle of Figure \ref{fig:model}, the representation (as key and value) does multi-head attention with the parameters of each prefix (as a query) corresponding to different initiatives. The output representations are then fed into a max-pooling layer \citep{dumoulin2016guide} and then fed into an MLP classifier to decide which initiative will guide the response generation.

\paragraph{Dialogue-Prefix Interaction.}
To recognize the potential initiative in the subsequent response generation, the initiative recognizer is implemented via a multi-head attention network. Let $\mathbf{h}_i\in\mathbb{R}^h$ be the contextualized representation of the $i$-th token in the sequence $S$, $\mathbf{h}^{I_t}_j\in\mathbb{R}^h$ be the $j$-th vector in the prefix query matrix $\mathbf{H}^{I_t}\in\mathbb{R}^{|\text{P}_{\text{idx}}|\times L\times h}$ of initiative $I_t$, where $\text{P}_{\text{idx}}$ denotes the sequence of prefix indices, and we use $|P_{\text{idx}}|$ to denote the length of the prefix. $L$ is the number of Transformer layers in a Transformer Decoder model (Decoder). Following \citet{li2021prefix}, $\mathbf{H}^{I_t}$ can be calculated by:
\begin{equation}
    \mathbf{H}^{I_t}=\text{MLP}_{\theta}(\mathbf{E}_\theta)~,
\end{equation}
where $\mathbf{E}_{\theta}\in\mathbb{R}^{|\text{P}_{\text{idx}}|\times h}$ is the embedding matrix of the prefix input tokens. Its parameter is tuned through the training process.

Then, the attention of the $\text{head}_\text{m}$  is calculated as:
\begin{equation}
e_{ij}=\frac{(\mathbf{h}_j^{I_t}\mathbf{W}^{Q})(\mathbf{h}_i\mathbf{W}^{K})^{\text{T}}}{\sqrt{d_z}}~,
\end{equation}
where $e_{ij}$ is the $\text{head}_\text{m}$ attention logit value between the $j$-th query vector of the prefix $I_t$ and the hidden vector of token $i$ in the dialogue history. $\sqrt{d_z}$ is a scaling factor \citep{vaswani2017attention}.

The attention scores can be calculated as:
\begin{equation}
\alpha_{ij}=\text{softmax}_i(e_{ij})=\frac{\text{exp}(e_{ij})}{\sum_{k\in\mathcal{N}}\text{exp}(e_{kj})}~,
\end{equation}
where $\mathcal{N}$ is the index set of the input sequence $S$.

Then, the  attention output in Figure \ref{fig:model} is calculated by:
\begin{equation}
\mathbf{z}_j^{\text{head}_m}=\sum_{i\in\mathcal{N}}\alpha_{ij}\mathbf{h}_i\mathbf{W}^{V}~,
\end{equation}
\begin{equation}
    \mathbf{z}_j=\text{Cat}\left[\mathbf{z}_j^{\text{head}_i},...,\mathbf{z}_j^{\text{head}_k}\right]~,
\end{equation}
where $\mathbf{W}^Q,\mathbf{W}^K,\mathbf{W}^V$ are the attention query, key, value matrices of $\text{head}_m$. $k$ is the number of attention heads, which is the same as that of in the Decoder. $\mathbf{z}_j^{\text{head}_m}$ is the output of the attention $\text{head}_m$. These head-based outputs are concatenated into the vector $\mathbf{z}_{j}$, which is then fused with the prefix query hidden representation $\mathbf{h}_j$:
\begin{equation}
    \mathbf{h}_{j}' = \text{MLP}(\mathbf{z}_{j})~,
\end{equation}
\begin{equation}
    \mathbf{h}_{j}'' = \text{Layernorm}(\mathbf{h}_{j}' + \mathbf{h}_{j})~,
\end{equation}
where ``MLP'' is a three-layer fully-connected neural network, and ``Layernorm'' is the Layer Normalization \cite{ba2016layer} operation.

All the outputs $[\mathbf{h}_1'',...,\mathbf{h}_{|\text{P}_{\text{idx}}|\times L}'']$ are organized into a matrix 
${\mathbf{H}^{I_t}}''\in\mathbb{R}^{|\text{P}_{\text{idx}}|\times L\times h}$
and then converted into a vector $\mathbf{h}^{I_t}\in\mathbb{R}^h$ for initiative classification via max-pooling \citep{dumoulin2016guide}:
\begin{equation}
    \mathbf{h}^{I_t}=\text{max-pooling}({\mathbf{H}^{I_t}}'')~.
\end{equation}

Finally, the initiative recognizer uses an MLP network to do initiative classification:
\begin{equation}
    \text{logit}_{I_t}=\text{MLP}(\mathbf{h}^{I_t})~,
\end{equation}
\begin{equation}
    P_{I_t}=\text{softmax}\frac{\text{exp}(\text{logit}_{I_t})}{\sum_{j=1}^{N}\text{exp}(\text{logit}_j)}~,
\end{equation}
where ``MLP'' is a three-layer fully-connected neural network, ``$P_{I_t}$'' is the probability of the initiative ${I_t}$.
Therefore, this classification loss can be:
\begin{equation}
    \mathcal{L}_{ce}=-\sum_{j=1}^{N}\text{log}(P_j)~.
\end{equation}

\subsubsection{Prefix-Based Response Generator}
To alleviate the cross-contamination problem in initiative-aware response generation, we propose two prefix-based response generation models: hard-attention generator and soft-attention generator. Different from the traditional attention mechanism, where attention is applied to input sequences (variables), the proposed two response generators impose the attention mechanism on \textit{prefix parameters}.

\paragraph{Soft-Attention Generator}
In the soft-attention generator, the prefix used in response generation derives from the soft-attention mechanism over all initiative-aware prefix parameters. Specifically, the prefix parameters for response generation are:
\begin{equation}
{\mathbf{H}^{I_t}}'=\sum_{j=1}^{N}P_{j}\mathbf{H}^{j}~.
\end{equation}

After that, this prefix is used to do response generation via a parameter-fixed Decoder model. Suppose the output matrix for the prefix is ${\mathbf{O}^{I_t}}'$:

\begin{equation}
  \mathbf{z}_i =
    \begin{cases}
      {\mathbf{O}^{I_t}}'[i,:] & \text{if $i\in \text{P}_{\text{idx}}$}~,\\
      \text{Decoder}_{\text{fixed}}(w_i,\mathbf{z}_{<i}) & \text{otherwise}~,
    \end{cases} 
    \label{eqn:prefix_ifelse}
\end{equation}
where $w_i$ is the $i$-th generated token.

Then, the distribution of the next token can be calculated by:
\begin{equation}
    p_{\psi}(w_{i+1}|\mathbf{z}_{\leq i})=\text{softmax}(\mathbf{W}_{\psi}^T\mathbf{z}_i)~,
    \label{eqn:token_pred}
\end{equation}
where $\mathbf{z}_i$ is the hidden states of token $w
_i$ in the last Transformers layer. $\mathbf{W}_{\psi}\in\mathbb{R}^{h\times|V|}$ is a matrix, mapping $\mathbf{z}_i$ to logits over vocabulary of size $|V|$.

Finally, the response generation loss is:
\begin{equation}
    \mathcal{L}_{gen}=\frac{1}{|Y_{t}|}\sum_{t=1}^{|Y_t|}p_{\psi}(Y_t|\mathbf{z}_{\leq t})~.
    \label{eqn:gen_loss}
\end{equation}

\paragraph{Hard-Attention Generator}
Different from the soft-attention generator, the hard-attention generator first selects the prefix ${\mathbf{H}^{I_t}}'\in\mathbb{R}^{2\times|\text{P}_{\text{idx}}|\times L \times h}$ with the highest initiative probability. 
Then, Equation \ref{eqn:prefix_ifelse}, \ref{eqn:token_pred} and \ref{eqn:gen_loss} are used to do response generation and calculate the generation loss.

\paragraph{Final Training Objective}
With all the above equations, the final training objective can be calculated as follows:
\begin{equation}
    \mathcal{L}=\gamma\cdot\mathcal{L}_{ce}+\alpha \cdot \mathcal{L}_{gen}~,
\end{equation}
where $\gamma, \alpha$ adjust the weight of losses. When $\gamma=1$, the Hard-Attention Generator is applied. When $\gamma=0$, the Soft-Attention Generator is applied.

\section{Experiments}
\subsection{Datasets}
We evaluate the proposed model IDPT on two dialogue datasets, the English version of DuRecDial v2.0 \citep{liu2021durecdial} and INSPIRED \citep{DBLP:conf/emnlp/HayatiKZSY20}. We annotated the initiatives of 700 turns of conversation for each dataset following the instruction in Appendix \ref{appendix:initiative_labeling} and had a substantial agreement (Cohen's Kappa 0.73 in DuRecDial v2.0 and 0.76 in INSPIRED). Then, for each dataset, we split the annotated turns into 500/100/100 for the training/dev/test set\footnote{In the following paper, ``DuRecDial v2.0'' or INSPIRED dataset denotes the 700 initiative-annotated data instances in this paper, ``training set''``dev set'' and ``test set'' refer to the specified 500, 100, and 100 data instances.}.

\subsection{Baselines}
We compare the proposed model IDPT with empathetic response generation models: MIME \citep{majumder2020mime} and EmpHi \citep{chen2022emphi}, end-to-end task-oriented dialogue model SimpleTOD \citep{hosseini2020simple}, promp-based methods: Manual Prompt Ensembling \citep{DBLP:conf/naacl/DevlinCLT19} and Prefix-Tuning \citep{li2021prefix},  initiative-aware knowledge-grounded dialogue model MIKe  \citep{meng2021initiative}, the constructed baseline Initiative-aware GPT-2s and the initiative-aware GPT-3.5 and GPT-4 \citep{DBLP:journals/corr/abs-2305-04147}. Details of baselines are in Appendix \ref{appendix:baselines}.

\begin{table*}[t]
\small
    \centering
    \begin{tabular}{lccc|ccc}    
        \toprule
        \multirow{2}{*}{Datasets} & \multicolumn{3}{c}{\textbf{DuRecDial v2.0}} & \multicolumn{3}{c}{\textbf{INSPIRED}}\\
        & Acc & BLEU-1/4 & ROUGE-1/L & Acc & BLEU-1/4 & ROUGE-1/L \\
        \midrule
        \multicolumn{7}{l}{\textit{100 examples}}\\
        \midrule
        EmpHi \citep{chen2022emphi}& 0.63 & 1.51/0.50 & 1.02/0.93 & 0.58 & 0.47/0.13 & 0.30/0.25\\
        MIME \citep{majumder2020mime}& 0.28 & 5.52/2.19 & 8.51/7.80 & 0.2 & 7.26/1.85 & 6.96/6.31\\
        SimpleTOD \citep{hosseini2020simple}& 0.62 & 26.00/9.36 & 19.24/17.68 & 0.72 & 19.00/4.80 & 11.19/9.36\\
        MIKe \citep{meng2021initiative}& 0.60 & 21.03/6.76 & 14.09/13.07 & 0.46 & 18.03/4.26 & 9.48/8.48\\
        Initiative-aware GPT-2s & 0.60 & 22.72/7.68 & 15.30/14.66 & 0.65 & 20.78/5.50 & 13.00/12.07\\
        Prefix Tuning  \citep{li2021prefix}& 0.60 & 18.44/9.07 & 16.61/16.24 & 0.65 & 18.60/5.31 & 13.52/12.24\\
        Manual Prompt Ensembling & 0.60 & 27.19/9.57 & 18.09/17.16 & 0.65 & {23.99/6.90} & 13.94/12.12\\
        GPT-3.5~\citep{DBLP:journals/corr/abs-2305-04147}  & 0.62 & 24.24/5.53 & 7.57/6.50 & 0.65 & 24.02/6.99 & 11.37/9.56\\
        GPT-4~\citep{DBLP:journals/corr/abs-2305-04147}  & 0.62 & \textbf{34.27}/8.50 & 14.12/11.59 & 0.65 & \textbf{34.57/7.53} & 12.51/9.57\\
        \hline
        IDPT-SA & 0.57$\dagger$ & 32.79/\uwave{13.34} & \uwave{26.92/25.69} & 0.62$\dagger$ & 20.62/5.68 & \uwave{14.78/13.36}\\
        IDPT-HA & \textbf{0.64} & \uwave{34.26}/\textbf{14.89} & \textbf{27.63/26.52} & \textbf{0.73} & \uwave{24.67/7.05} & \textbf{15.87/13.74}\\
        \midrule
        \multicolumn{7}{l}{\textit{500 examples}}\\
        \midrule
        EmpHi \citep{chen2022emphi}& 0.66 & 3.16/1.05 & 3.02/2.97 & 0.63 & 3.08/0.84 & 2.11/2.09\\
        MIME \citep{majumder2020mime}& 0.63 & 26.84/6.22 & 13.57/13.00 & 0.73 & 25.13/5.03 & 10.49/8.74\\
        SimpleTOD \citep{hosseini2020simple}& 0.82 & 37.52/18.74 & 32.21/30.55 & 0.76 & \uwave{29.03/7.85} & 15.31/13.21\\
        MIKe \citep{meng2021initiative}& 0.86 & 33.95/16.21 & 26.05/25.32 & 0.68 & 21.15/4.85 & 9.02/7.67\\
        Initiative-aware GPT-2s& 0.76 & 32.63/13.89 & 24.33/23.45 & 0.75 & 26.51/7.61 & 15.78/14.04\\
        Prefix Tuning \citep{li2021prefix}& 0.76 & 31.69/15.95 & 29.35/27.80 & 0.75 & 23.21/8.31 & 17.12/15.43\\
        Manual Prompt Ensembling & 0.76 & 38.30/19.22 & 31.36/30.37 & 0.75 & 26.88/8.16 & 15.37/13.63\\
        \hline
        IDPT-SA & 0.77$\dagger$ & \uwave{44.48/22.45} & \uwave{37.55/36.08} & 0.78$\dagger$ & 28.10/9.59 & \uwave{18.89/17.42}\\
        IDPT-HA & \textbf{0.90} & \textbf{46.50/24.35} & \textbf{39.56/38.01} & \textbf{0.85} & \textbf{30.13/10.31} & \textbf{20.93/19.02}\\
        \bottomrule   
    \end{tabular}
\caption{\label{tab:automatic_main_result1}Automatic evaluation results on DuRecDial v2.0 and INSPIRED with training sizes 100, and 500.``IDPT-HA'' uses the hard-attention generator while ``IDPT-SA'' uses the soft-attention generator.  The best results are in bold font and the second best results are underlined with wave line. $\dagger$ denotes the performance without initiative labels but calculated by the unsupervised initiative logits.}
\end{table*}

\subsection{Evaluation Metrics}
For automatic evaluation,we calculate the prediction accuracy of initiative classification in each model, and we evaluate the generated responses with BLEU-1/4 \citep{DBLP:conf/acl/PapineniRWZ02} and ROUGE-1/L \citep{lin2004rouge}.

We conduct our human evaluation by hiring three well-educated annotators. We provide the  dialogue history, the response generated by IDPT as well as the response generated by another method (their names are marked out). Each annotator needs to independently annotate their preference for each pair of responses based on three aspects: readability, informativeness, and engagingness. The results are reported in Table \ref{table:ab_test}.

\subsection{Experimental Settings}
In the proposed IDPT, we take BERT model~\citep{DBLP:conf/naacl/DevlinCLT19} as the Encoder and the GPT-2 model~\citep{radfordlanguage} as the decoder\footnote{We also tried T5~\citep{raffel2020exploring} for initiative prediction and response generation in a single model. However, the performance is poor.}. To evaluate the performance of the compared methods on initiative classification and response generation, we vary the training size from 100 to 500 examples, except for GPT-3.5 and GPT-4~\citep{DBLP:journals/corr/abs-2305-04147}. The reason is that both models have a performance plateau when the training size is larger than 100. More details about model implementation and training details are in Appendix \ref{sec:implement}.

\begin{table*}[!htp]
\fontsize{10}{11}\selectfont
    \centering
    \begin{tabular}{lcccccccccccccccccc}    
        \toprule
        \multirow{2}{*}{Methods} & \multicolumn{9}{c}{\textbf{DuRecDial v2.0}} \\\cline{2-10}
        & \multicolumn{3}{c}{Readability} &\multicolumn{3}{c}{Informativeness} & \multicolumn{3}{c}{Engagingness}\\
        & Win & Tie & Lose & Win & Tie & Loss& Win & Tie & Lose \\
        \midrule
        IDPT v.s. Prefix-Tuning & 30&49&21&50&20&30& 46&30&24\\
       IDPT v.s. SimpleTOD & 33&	46&	 21&40&	40&	20& 51&25&24\\
        IDPT v.s. Manual Prompt Ensembling & 24&	54&	22&	37&	45&	18& 30&46&24\\
        IDPT v.s. GPT-4 & 20&	64&	16&	35&	47&	18& 24&56&20\\
        \bottomrule   
    \end{tabular}
    \caption{\label{table:ab_test}Human evaluation on the test set of DuRecDial v2.0. The result of INSPIRED is in Appendix \ref{appendix:a_b_test_inspired}.} 
    \vspace{+0.2cm}
\end{table*}

\begin{table*}[!htp]
\fontsize{10}{11}\selectfont
    \centering
    \begin{tabular}{lccc|ccc}    
        \toprule
        \multirow{2}{*}{Datasets} & \multicolumn{3}{c}{\textbf{DuRecDial v2.0}} & \multicolumn{3}{c}{\textbf{INSPIRED}}\\
        & Acc & BLEU-1/4 & ROUGE-1/L & Acc & BLEU-1/4 & ROUGE-1/L \\
        \midrule
        IDPT & \textbf{0.83}& \textbf{36.70/17.87}&\textbf{34.11/31.73}& \textbf{0.67}&\textbf{28.27/11.83}&	\textbf{21.07/19.72}\\
        w/ Averaging Prefixes & 0.56&	28.22/12.99&	22.76/21.52&	0.65&	13.01/3.37&	10.91/9.52\\
        w/ Randomly Selected Prefix & 0.57&	33.27/17.06&	31.49/29.57&	0.65&	26.83/10.04&	19.85/18.55\\
        w/o Initiative & -&	29.79/14.05&	26.61/25.44&	-&	23.40/11.74&	19.09/18.58\\
        \bottomrule   
    \end{tabular}
    \caption{\label{tab:ablation}Ablation study of IDPT on the dev set of DuRecDial v2.0 and INSPIRED. ``w/ Averaging Prefixes'' denotes that prefix parameters of all initiatives are averaged into a single prefix. ``w/ Randomly Selected Prefix'' is that the prefix associated with a specific initiative label is randomly selected. In ``w/o'' Initiative'', the initiative classifier is removed and only prefix tuning is used for response generation.} 
    \vspace{-0.2cm}
\end{table*}
\subsection{Main Results}

For automatic evaluation, Table \ref{tab:automatic_main_result1} (more results in Appendix \ref{appendix:extra_main_results}) shows that IDPT-HA and IDPT-SA outperform all baselines on generation metrics (BLEU-1/4, ROUGE-1/L) with 100 to 500 labeled instances. GPT-3.5 and GPT-4 have high BLEU-1, as they generate more informative responses, but low BLEU-4 and ROUGE-1/L, indicating the lack of initiative ability in LLM and the strength of IDPT. ``Initiative-aware GPT-2s'' are worse than IDPT and some baselines, despite using much more disk space (about 2000x larger). IDPT-HA is better than IDPT-SA, and the second best results in BLEU-1/4 for INSPIRED are not IDPT-SA, suggesting the importance of initiative supervision. Hence, we use IDPT-HA for further analysis.

For human evaluation, as shown in Table \ref{table:ab_test}, we conduct an A/B test among the proposed IDPT and four strong baselines: Prefix-Tuning, SimpleTOD, Manual Prompt Ensembling, and GPT-4. Overall, the proposed IDPT achieves the best performance on each dataset, especially on informativeness. For example, the win ratio of the proposed IDPT v.s. the most competitive baseline GPT-4 is 35\% on DuRecDial v2.0 and 25\% on INSPIRED, which can be explained that initiative-related information proposed in this work can guide the system to provide more information.

\subsection{Detailed Analysis}
We further analyze the effectiveness of the proposed model trained on 500 instances in detail.
\subsubsection{Ablation Study}

We ablate three components of IDPT to evaluate their effectiveness. Table \ref{tab:ablation} shows that, first, averaging the prefix parameters (``w/ Averaging Prefixes'') lowers the performance, as it confuses the generation model and prevents it from producing the correct initiative-aware response. Second, randomly selecting the prefix (``w/ Random Selected Prefix'') also performs worse than IDPT, as it causes a mismatch between the desired initiative and the wrong prefix label, indicating the importance of a good initiative classifier. Third, removing the initiatives (``w/o Initiatives'') reduces the performance, demonstrating the value of the initiative signal in response generation.

\begin{figure}[!htp]
\vspace{-0.2cm}
\centering
\subfloat[DuRecDial V2.0]{%
  \includegraphics[clip,width=.56\columnwidth]{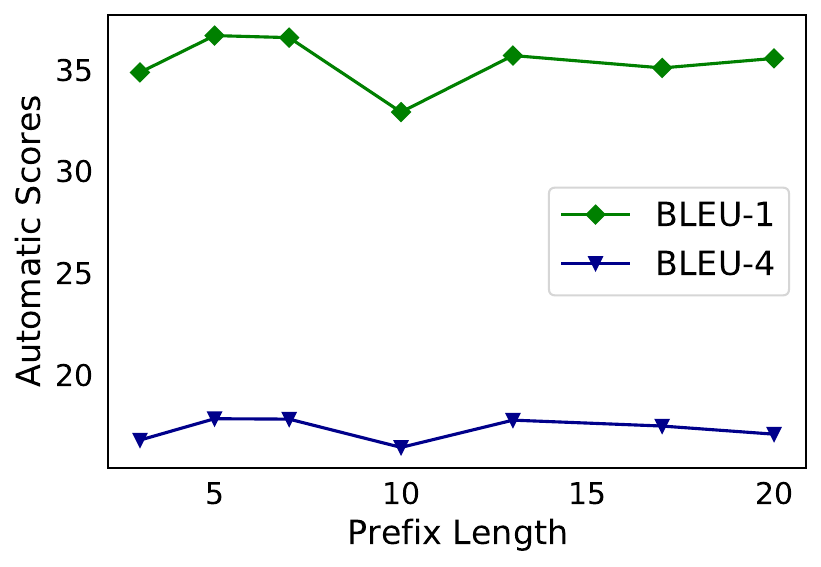}%
}\\
\subfloat[INSPIRED]{%
  \includegraphics[clip,width=.56\columnwidth]{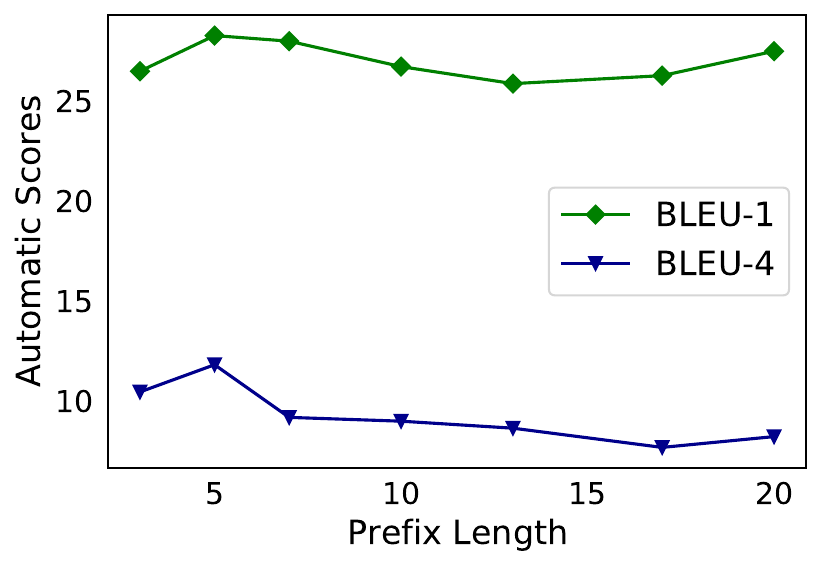}%
}
\vspace{-0.3cm}
\caption{\label{fig:prefix_len}BLEU-1/4 scores of the proposed IDPT with the varied prefix length.}
\vspace{-0.1cm}
\end{figure}

\begin{table*}[!htp]
\small
  \centering
  \begin{tabular}{p{.47\textwidth}|p{.47\textwidth}}
    \toprule
    \textbf{Context}: &\textbf{Context}:\\
    System: How's life recently? & System: Santa Paws is a very good movie.\\
    User: I'm a little tired looking after my grandson these days. & User: Oh that sounds good. Is it kid friendly?\\
    \midrule
    \textbf{Initiative Type}: dialogue-level: system, utterance-level: system&\textbf{Initiative Type}: dialogue-level: system, utterance-level: user\\
    \midrule
    \textbf{Response}:&\textbf{Response}:\\
    \textbf{Prompt Ensembling}: It's very easy to get to know the leading actress of this song.&\textbf{Prompt Ensembling}: it's a little bit of both.\\
    \textbf{SimpleTOD}: They're great hobbies.&\textbf{SimpleTOD}: Yes, I am looking for a movie trailer with a good message to the movie.\\
    \textbf{IDPT}: You can relax and listen to some songs. what's your favorite song? &\textbf{IDPT}: Yes it is!\\
    \textbf{Reference Answer}: Listen to some music and relax a bit. I'd recommend the song Families to you.&\textbf{Reference Answer}: Its plenty of kid friendly.\\
    \bottomrule
  \end{tabular}
  \caption{\label{tab:case_study_with_baselines}Case study among IDPT and other baseline methods. The two examples are from the test set of DuRecDial v2.0 (left hand side) and INSPIRED (right hand side).}
  \vspace{-0.2cm}
\end{table*}

\subsubsection{Effects of Prefix Length}

We examine the effects of prefix length on IDPT's performance. Figure \ref{fig:prefix_len} shows the improvement from length 3 to 5 for both DuRecDial V2.0 and INSPIRED datasets. Then, a slight decrease from 7 to 20, especially for INSPIRED, due to overfitting in the low-data setting. Therefore, we choose ``5'' as the optimal prefix length for both datasets.

\begin{table}[!htp]
\small
  \centering
  \begin{tabular}{p{.45\textwidth}}
    \toprule
    \textbf{Context}: \\
    User: Please play the song Thousand Sad Reasons for me. \\
    System: The Thousand Sad Reasons is on. Please enjoy.\\
    User: Okay, well done! Pretty nice song!\\
    \midrule
    \textbf{Initiative Type}: dialogue-level: user, utterance-level: user (initiative 0)\\

    \textbf{Response}: That's my idol's song.\\
    \midrule
    \textbf{Initiative Type}: dialogue-level: user, utterance-level: system (initiative 1)\\

    \textbf{Response}: The song is really classic.\\
    \midrule
    \textbf{Initiative Type}: dialogue-level: system, utterance-level: user (initiative 2)\\

    \textbf{Response}: I've heard this song. I'd like to recommend another one to you!\\
    \midrule
    \textbf{Initiative Type}: dialogue-level: system, utterance-level: system (initiative 3)\\

    \textbf{Response}: I'd like to recommend another song Be Torn With Grief by Jacky Cheung. It's a classic lyrical pop song.\\
    \bottomrule
  \end{tabular}
  \caption{\label{tab:case_study_given_diff_init}Case study of IDPT with different initiatives. This example is from the test set of DuRecDial v2.0.}
\end{table}

\subsubsection{Case Study}
We present two case studies in this section to illustrate the benefits of IDPT. As Table \ref{tab:case_study_with_baselines} shows, IDPT can leverage the initiative labels more effectively for response generation. For instance, in the left example, when both initiatives are system, the system should initiate a new topic. The manual prompt ensembling and SimpleTOD responses are off-topic, while IDPT proactively suggests a song and transitions smoothly from ``tired'' to ``relax''. 
For the example on the left of Table \ref{tab:case_study_with_baselines}, given the initiative type: dialogue-level system and utterance-level user, it means to keep the high-level topic held by the system (movie ``Santa Paws'') and following the sub-topic (``friendly'' or not) drawn by the user. Only IDPT generates a proper response with the given initiative label.

We also demonstrate an example of generating different responses with varied initiatives. Table \ref{tab:case_study_given_diff_init} shows that, given initiative 0, the system follows the user by saying, ``That's my idol's song.'' Given initiative 1, the system introduces a new sub-topic (``classic'' style) within the main topic ``music''. Given initiative 2, the system partially changes the main topic to another music. Given initiative 3, the system completely changes the direction by suggesting a new song ``Be Torn With Grief''. This shows the model's controllability over response generation.

\section{Conclusion}
In this research, we underscored the significance of considering initiative when generating responses in dialogue systems. To this end, we introduced the Mix-Initiative Dynamic Prefix Tuning (IDPT) model, which effectively incorporates initiative cues by segregating initiative elements into distinct prefix parameters. Recognizing the challenges posed by the scarcity of labeled initiative conversation data, our model is adept at adapting to both supervised and unsupervised learning environments. Comprehensive experiments on two widely-used dialogue datasets affirms the superior performance of the proposed IDPT model as compared to a series of baselines. Notably, our proposed IDPT model consistently produces suitable responses for the conversation context even when initiatives are intentionally manipulated.

\section*{Limitations}
Our work has the following limitations. Firstly, there is still room to further improve the initiative classification performance as well as the response generation performance. Recently, Large Language Models (LLMs) such as GPT-3.5 and GPT-4 have shown superior performance on various tasks, thanks to the abundant semantic knowledge embedded in the large scale model parameters.
Although our model has the advantage of flexibility and relatively low cost, it might also be useful to leverage the abundant prior knowledge in such models, especially under the supervised and unsupervised setting. We will further explore this possibility. Secondly, although the proposed IDPT has shown superior performance under the supervised learning setting, it is unclear how it will perform when large number of initiative annotated dialogues is available. We will look into it in the future. 
\section*{Ethics Statement}

In our pursuit to augment the initiative-awareness of dialogue systems with IDPT, thereby aiming to heighten user satisfaction, several ethical implications have emerged that need careful consideration.

One issue is the concern of safety and the potential malevolent impact of the initiative-driven agent \cite{deng2023prompting}. If the agent is nurtured on a compromised dataset, it risks churning out detrimental responses, which may be deleterious for users. This concern escalates when implemented in sensitive domains like healthcare or finance, where misguidance can jeopardize user well-being or financial interests. As a safeguard, it becomes imperative to vouch for the quality of the training data, and to continually appraise the agent across diverse contexts.

Another issue is the transparency and accountability of the agent. Since IDPT uses dynamic prefix tuning to adjust the initiative, it may be hard for the user to know why and how the agent chooses an initiative. This could affect the user's trust and confidence in the agent, especially if the agent makes an error or a controversial decision. Moreover, it may raise questions about who is liable for the outcomes of the agent's actions, such as providing false information or influencing the user's opinion. Therefore, it is essential to provide clear and explicit explanations for the agent's initiative choices, and to establish appropriate mechanisms for feedback and correction.

\section*{Acknowledgements}
The work is supported by MIIT Program (CEIEC-2022-ZM02-0247), National Natural Science Foundation of China (No. U21B2009, 62172039 and 62276110). We thank the ACL ARR reviewers for their helpful feedback. We would like to acknowledge Sihan Guo, Shuhang Liu, Youchao Zhou, and others for the helpful discussions.

\bibliography{anthology,custom}

\appendix
\clearpage
\newpage

\section{Initiative Labeling}
\label{appendix:initiative_labeling}

\numberwithin{equation}{section}
\numberwithin{figure}{section}
\numberwithin{table}{section}
\begin{table}[!htp] 
    \centering
    \begin{tabular}{ccc}    
        \hline
         &DuRecDial v2.0& INSPIRED\\
         \hline
        Initiative 0& 169 & 39\\
        Initiative 1& 20 & 42\\
        Initiative 2& 80 & 141\\
        Initiative 3& 431 & 478\\
        \hline
    \end{tabular}
    \caption{\label{tab:stat_init_labels} Statistics of annotated initiative labels in DuRecDial v2.0 and INSPIRED.} 
\end{table}

We hired three well-educated people to independently label each response. Thus, each dialogue instance is a triple (dialogue history, response, initiative). Finally, 700 turns of responses are annotated with initiatives for each dialogue dataset. The final initiative labels are selected based on the majority votes among the three annotated copies. Otherwise, we randomly selected one initiative label from the three different annotated labels.

The statistics of initiative labels are in Table \ref{tab:stat_init_labels}. It shows that labels are unbalanced in these two datasets, where initiative 3 makes up a large part of the entire dataset. Thus, a single response generation model could probably lean toward responding quite proactively (initiative 3), further causing a severe cross-contamination problem. 

The annotated initiative-aware responses will be released for academic usage (which is also known among the hired annotators) upon acceptance.

\section{Main Results of 200 and 400 Training Sizes}
\label{appendix:extra_main_results}

\numberwithin{equation}{section}
\numberwithin{figure}{section}
\numberwithin{table}{section}
For the sake of page limit, we report the automatic results of 200 and 400 training sizes in Table \ref{tab:automatci_main_reult_2}.

\begin{table*}[!htp]
\small
    \centering
    \renewcommand*{\arraystretch}{1.2}
    \begin{tabular}{lcccccc}    
        \toprule
        \multirow{2}{*}{Datasets} & \multicolumn{3}{c}{DuRecDial v2.0} & \multicolumn{3}{c}{INSPIRED}\\
        & Acc & BLEU-1/4 & ROUGE-1/L & Acc & BLEU-1/4 & ROUGE-1/L \\
        \midrule
        \multicolumn{7}{l}{\textit{200 examples}}\\
        \midrule
        EmpHi \citep{chen2022emphi}& 0.64 & 1.85/0.47 & 1.46/1.32 & 0.58 & 2.19/0.82 & 2.03/1.88 \\
        MIME \citep{majumder2020mime}& 0.63 & 17.26/4.58 & 9.72/8.74& 0.73 & 9.41/2.61 & 8.10/7.13\\
        SimpleTOD \citep{hosseini2020simple}& 0.70 & 32.06/13.12 & 22.35/21.20 & 0.72 & 24.70/6.40 & 13.87/12.08\\
        MIKe \citep{meng2021initiative}& 0.83 & 27.43/9.88 & 18.97/17.75 & 0.5 & 20.23/4.65 & 9.19/8.30\\
        Initiative-aware GPT-2s & 0.63 & 32.67/12.91 & 24.64/22.93 & 0.70 & 23.74/5.63 & 12.91/11.27\\
        Prefix Tuning  \citep{li2021prefix}& 0.63 & 23.53/9.90 & 19.92/18.31 & 0.70 & 20.10/6.35 & 14.15/12.78\\
        Manual Prompt Ensembling & 0.63 & 33.26/13.92 & 25.70/24.33 & 0.70 & 24.66/15.50 & 12.99/11.68\\
        \hline
        IDPT-SA & 0.60$\dagger$ & \uwave{37.95/18.39} & \uwave{32.20/30.59} & 0.67$\dagger$ & \uwave{25.06/7.53} & \uwave{18.24/15.78}\\
        IDPT-HA & \textbf{0.85} & \textbf{39.32/20.29} & \textbf{33.20/32.45} &\textbf{0.75} & \textbf{26.09/8.60} & \textbf{19.02/17.20}\\
        \midrule
        \multicolumn{7}{l}{\textit{300 examples}}\\
        \midrule
        EmpHi \citep{chen2022emphi}& 0.64 & 2.80/1.31 & 2.54/2.38 & 0.60 & 2.33/0.65 & 2.06/1.95\\
        MIME \citep{majumder2020mime}& 0.63 & 23.87/6.62 & 9.13/8.97 & 0.73 & 21.33/5.03 & 8.31/6.68 \\
        SimpleTOD \citep{hosseini2020simple}& 0.78 & 35.48/16.23 & 28.90/27.35 & 0.73 & 25.87/6.97 & 14.82/12.68\\
        MIKe \citep{meng2021initiative}& 0.83 & 28.99/11.56 & 21.53/20.83 & 0.54 & 20.71/4.74 & 9.47/8.11\\
        Initiative-aware GPT-2s & 0.67 & 34.50/15.57 & 26.97/25.72 & 0.72 & 23.75/5.82 & 13.31/11.83\\
        Prefix Tuning  \citep{li2021prefix}& 0.67 & 23.66/16.57 & 16.60/15.22 & 0.72 & 20.89/6.85 & 14.92/14.01\\
        Manual Prompt Ensembling & 0.67 & 35.56/16.01& 30.16/29.15 & 0.72 & \uwave{26.92/6.61} & 15.37/13.63\\
        \hline
        IDPT-SA & 0.63$\dagger$ & \uwave{42.73/20.25} & \uwave{36.17/34.19} & 0.73$\dagger$ & 26.75/7.93 & \uwave{18.92/16.76}\\
        IDPT-HA & \textbf{0.88} & \textbf{43.02/22.49} & \textbf{37.60/36.09} & \textbf{0.78} & \textbf{28.09/8.34} & \textbf{20.62/18.30}\\
        \midrule
        \multicolumn{7}{l}{\textit{400 examples}}\\
        \midrule
        EmpHi \citep{chen2022emphi}& 0.65 & 2.98/0.49 & 2.55/2.23 & 0.62 & 2.38/0.66 & 2.12/1.99\\
        MIME \citep{majumder2020mime}& 0.63 & 23.89/5.42 & 11.53/10.44 & 0.73 & 22.76/5.10 & 9.54/8.06 \\
        SimpleTOD \citep{hosseini2020simple}& 0.80 & 36.61/17.50 & 30.09/29.34 & 0.73 & 26.98/7.32 & 14.63/12.96\\
        MIKe \citep{meng2021initiative}& 0.84 & 33.15/14.68 & 24.55/23.31 & 0.62 & 21.14/4.77 & 10.16/8.96\\
        Initiative-aware GPT-2s & 0.73 & 36.55/16.29 & 28.27/27.06 & 0.74 & 24.10/6.28 & 13.87/12.13\\
        Prefix Tuning  \citep{li2021prefix}& 0.73 & 31.73/15.91 & 28.14/16.89 & 0.74 & 21.18/6.37 & 14.57/13.41\\
        Manual Prompt Ensembling & 0.73 & 38.16/17.49 & 31.46/30.12 & 0.74 & 27.07/7.96 & 14.88/12.92\\
        \hline
        IDPT-SA & 0.71$\dagger$ & \uwave{42.86/20.75} & \uwave{37.35/35.84} & 0.73$\dagger$ & \uwave{27.72/10.06} & \uwave{18.29/17.03}\\
        IDPT-HA & \textbf{0.88} & \textbf{44.47/22.74} & \textbf{39.05/37.35} & \textbf{0.82} & \textbf{29.07/10.41} & \textbf{18.96/17.14}\\
        \bottomrule   
    \end{tabular}
    \caption{\label{tab:automatci_main_reult_2}Automatic evaluation results on DuRecDial v2.0 and INSPIRED with training sizes 200, 300, and 400.} 
\end{table*}

\section{Human Evaluation Results on INSPIRED}
\label{appendix:a_b_test_inspired}
The human evaluation results on INSIPRED is shown in Table \ref{table:ab_test2}, from which it is known that responses generated by IDPT is better especially in ``Informativeness'' and ``Engagingness''. It is consistent with the effects of initiatives, which can help the system to actively join the conversation and provide more useful information.
\numberwithin{equation}{section}
\numberwithin{figure}{section}
\numberwithin{table}{section}
\begin{table*}[t]
\small
    \centering
    \renewcommand*{\arraystretch}{1.2}
    \begin{tabular}{lccccccccc}    
        \toprule
        \multirow{3}{*}{Methods} & \multicolumn{9}{c}{INSPIRED} \\\cline{2-10}
        & \multicolumn{3}{c}{Readability} &\multicolumn{3}{c}{Informativeness} & \multicolumn{3}{c}{Engagingness}\\
        & Win & Tie & Lose & Win & Tie & Loss& Win & Tie & Lose \\
        \midrule
        IDPT v.s. Prefix-Tuning & 42&40&18&67&28&5& 52&30&18\\
       IDPT v.s. SimpleTOD & 41&	23&	 36&26&	59&	15& 52&38&10\\
        IDPT v.s. Manual Prompt Ensembling & 20&	62&	18&	26&	59&	15& 34&44&22\\
        IDPT v.s. GPT-4 & 20&	60&	20&	25&	58&	17& 30&42&28\\
        \bottomrule   
    \end{tabular}
    \caption{\label{table:ab_test2}Human evaluation on the test set of INSPIRED.} 
\end{table*}

\section{Exploring the Relationship Between Response Length and Initiative Labels}

In our examination of the potential correlation between response length and initiative labels, we employed the Pearson correlation coefficient\footnote{\url{https://en.wikipedia.org/wiki/Pearson_correlation_coefficient}} as a statistical measure. This analysis was conducted on two distinct datasets: the annotated DuRecDial and INSPIRED. The computed coefficients were 0.1263 and -0.0122, respectively. These findings indicate a negligible correlation between the length of responses and the initiative labels. Consequently, it can be inferred that response length does not significantly influence the model's ability to differentiate among various initiatives.

\section{Baselines in Details}
\label{appendix:baselines}
{\setlength{\parindent}{0cm}
\textbf{MIME} \citep{majumder2020mime} aims to generate empathetic responses by recognizing and mimicking varied emotions from human utterances. In this work, we replace ``emotion'' by ``initiative'' to let initiative labels control the response generation.
}

{\setlength{\parindent}{0cm}
\textbf{SimpleTOD} \citep{hosseini2020simple} unify all subtasks of a task-oriented system into a sequence prediction task. For initiative-aware response generation, given dialogue history, we first let the model generate initiative labels (e.g. text ``1, 1'' corresponding initiative label 3), followed by response generation.
}

{\setlength{\parindent}{0cm}
\textbf{Manual Prompt Ensembling.} We first train a BERT model \citep{DBLP:conf/naacl/DevlinCLT19} as the initiative classifier. Then, we design three different initiative-aware manual templates to guide a GPT-2 model and ensemble them for response generation. For example, given the dialogue history and an initiative, the format of prompt 1 is: \texttt{[Dialogue History] dialogue initiative: [DI] utterance initiative: [UI] [Response]}. The format of prompt 2 is: \texttt{[Dialogue History] Please use the dialogue initiative:[DI\_ROLE] and utterance initiative [UI\_ROLE] initiative to give a response: [Response]}; and the format of prompt 3 is: \texttt{[Dialogue History] [DI], [UI] [Response]}. Here, ``DI'' and ``UI'' denote the binary values of dialogue-level and utterance-level initiative. ``DI\_ROLE'' and ``UI\_ROLE'' are text forms of the aforementioned binary values. For example, if DI=1, DI\_ROLE is ``user'', otherwise ``system''. For each template, the initiative slot is pre-filled with the predicted result from the initiative classifier. For response generation, we ensemble these three prompt-based methods by averaging the probabilities of each predicted token.
}

{\setlength{\parindent}{0cm}
\textbf{Initiative-aware GPT-2s .} In this baseline, a GPT-2 model is fine-tuned only on the response subset with a specific initiative. In the inference time, a BERT-based initiative classifier is applied to decide which initiative would be selected. Then, the trained GPT-2 model with the selected initiative is used to do response generation.
}

{\setlength{\parindent}{0cm}
\textbf{Prefix-Tuning} \citep{li2021prefix} is a prompted-based text generation method. With a tunable task-specific prefix, it can generate desired and task-related text. To incorporate initiatives into it, we separately train a prefix corresponding to each initiative and uses a BERT-base-uncased model as an initiative classifier to decide which prefix to choose in the generation step.
}

{\setlength{\parindent}{0cm}
\textbf{MIKe} \citep{meng2021initiative} propose a self-supervised initiative-aware method to improve the performance of knowledge selection in the knowledge-grounded dialogue.  We keep the teacher-student initiative recognition module and use the truth initiative labels to calculate the initiative prediction loss while jointly training a response generation model.
}

{\setlength{\parindent}{0cm}
\textbf{EmpHi} \citep{chen2022emphi} is to recognize emotions as well as intents in the dialogue for response generation. We replace the emotion prediction part with an initiative prediction part. 
}

{\setlength{\parindent}{0cm}
\textbf{Initiative-aware GPT-3.5/4} \citep{DBLP:journals/corr/abs-2305-04147} is to prompt responses with specific initiative. We use the similar manual prompt, where the background and the training examples are given to the GPT-3.5/4 model to make it familiar with the task. And we want the model to generate the response given specific dialogue history as well as the initiative provided by a BERT classifier. 
}

\section{Implementation Details}
\label{sec:implement}
For all but EmpHi, we use the base version of GPT-2 as the response generation backbone model. For the manual prompt ensembling, initiative-aware GPT-2s and prefix-tuning baselines, we use BERT-base-uncased as the backbone of the initiative classifier. The prefix length in prefix-tuning is 10. In the proposed IDPT, for each prefix, the length is 5, and the adjusting weight $\alpha=1$. We searched over \{5e-6,6e-4,7e-6,8e-6,9e-6,1e-4\} learning rate, \{10, 10, 50, 100, 150\} training epochs, and \{1, 2, 4, 8, 16\} batch size. The final learning rate is 8e-5, the epoch number is 100 and the training batch size is 4. It took 2.2 hours to train the model on a single GPU with 11GB memory.

\section{Extra Information of Datasets}
Both the DuRecDial v2.0 and INSPIRED are created for academic usage, which is consistent with the purpose of this paper.

The License of DuRecDial v2.0 is Apache-2.0 license while there is not a license for INSPIRED.

\end{document}